\documentclass[letterpaper, 10 pt, conference]{ieeeconf}  

\usepackage{tikz}
\usepackage{algpseudocode}
\usepackage{algorithm}
\usepackage{siunitx}
\usepackage{pgfplots}
\usepackage{wrapfig}
\usepackage{placeins}
\usepackage{makecell}

\usepackage{booktabs}

\usepackage{amssymb}
\usepackage{amsmath}

\usepackage{caption}
\usepackage{subcaption}
\usepackage{dsfont}

\usepackage{upgreek}
\pgfplotsset{compat=1.5.1}

\IEEEoverridecommandlockouts                              

\title{\LARGE \bf
Differentiable Constrained Imitation Learning for Robot Motion Planning and Control
}

\author{Christopher Diehl, Janis Adamek, Martin Kr\"uger, Frank Hoffmann and Torsten Bertram
\thanks{This research was funded by the Federal Ministry for Economic Affairs and Climate Actions on the basis of a decision by the German Bundestag and the European Union in the project "KISSaF - AI-based Situation Interpretation for Automated Driving".}
\thanks{The authors are with the Institute of Control Theory and Systems Engineering, TU Dortmund University, D-44227, Germany.}%
}

\begin{document}

\maketitle
\thispagestyle{empty}
\pagestyle{empty}

\begin{abstract}
Motion planning and control are crucial components of robotics applications like automated driving. Here, spatio-temporal hard constraints like system dynamics and safety boundaries (e.g., obstacles) restrict the robot's motions. 
Direct methods from optimal control  solve a constrained optimization problem. However, in many applications finding a proper cost function is inherently difficult because of the weighting of partially conflicting objectives. On the other hand, Imitation Learning (IL) methods such as Behavior Cloning (BC) provide an intuitive framework for learning decision-making from offline demonstrations and constitute a promising avenue for planning and control in complex robot applications. Prior work primarily relied on soft constraint approaches, which use additional auxiliary loss terms describing the constraints. However, catastrophic safety-critical failures might occur in out-of-distribution (OOD) scenarios. This work integrates the flexibility of IL with hard constraint handling in optimal control. Our approach constitutes a general framework for constraint robotic motion planning and control, as well as traffic agent simulation, whereas we focus on mobile robot and automated driving applications. Hard constraints are integrated into the learning problem in a differentiable manner, via explicit completion and gradient-based correction. Simulated experiments of mobile robot navigation and automated driving provide evidence for the performance of the proposed method.
\end{abstract}
%
\section{Introduction}
The motion of robots in the real world is constrained by the kinematics and dynamics of the robot as well as the geometric structure of the environment. For example, to navigate safely and smoothly, a self-driving vehicle (SDV) must consider various factors such as its control limits, stop signs, and obstacles building a driving corridor.  A core challenge is incorporating these constraints into robot planning and control. That is also essential for automated driving traffic simulation to enhance the realism of the simulated agents. For instance, traffic agents must follow common road rules. On the one side, optimal control approaches solve a finite horizon optimal control problem by optimizing a cost function under \textit{explicitly} defined constraints. A common approach, like in direct methods \cite{2010_John}, is to derive a nonlinear program from a continuous optimal control formulation \cite{Roesmann2017, Roesmann2021, Diehl2022} and then solve the problem with numerical optimization. However, designing a general cost function remains an unsolved problem for inherently complex tasks such as automated driving  \cite{Rewardmisdesign, diehl2021umbrella, RALDiehl}. Here, aspects like comfort and safety must be weighed against each other. On the other side, robot behavior can be learned from demonstrations, which is the task of IL. One example is BC, a simple \textit{offline} learning method, requiring no \textit{on-policy} environment interactions. Here, constraints are \textit{implicitly} learned from data. Further, constraints can be integrated by auxiliary loss functions. However, there are no guarantees for constraint satisfaction, and robot policies fail under distribution shifts \cite{Dagger}, causing unexpected unsafe actions.

\begin{figure}[!t]
	\centering
	
	\includegraphics[width=\columnwidth]{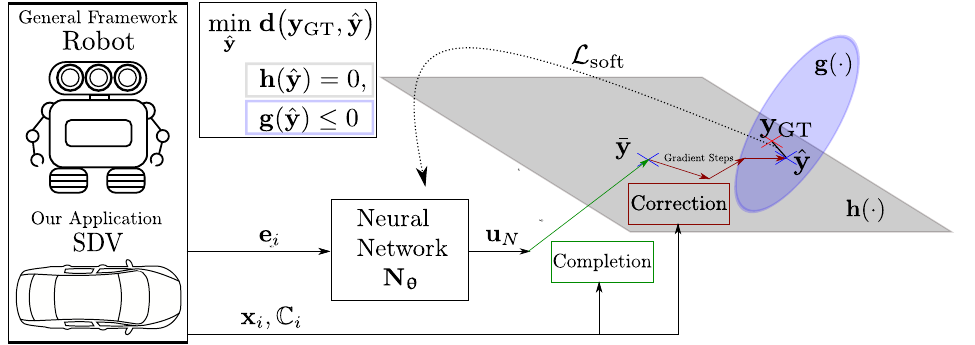}

	\caption{A schematic overview of the proposed framework: A robot, like an SDV, perceives its environment and builds a high-dimensional environment model $\mathbf{e}_i$ and a low-dimensional state representation $\mathbf{x}_i$. Constraints $\mathbb{C}_i$ (grey rectangle: equality constraints, blue ellipse: inequality constraints) further bound the robots motion. A neural network $\mathbf{N}_{\boldsymbol{\theta}}$ processes $\mathbf{e}_i$ and outputs an initial sequence of control values $\mathbf{u}_N$. These are \textit{completed} to the initial solution $\bar{\mathbf{y}}$, also containing the predicted states, by unrolling a robot dynamics model. Afterward, $\bar{\mathbf{y}}$ is \textit{corrected} with gradient steps (red arrows), such that the estimated solution $\hat{\mathbf{y}}$ lies in the space defined by equality (grey) and the inequality constraints (blue) of $\mathbb{C}_i$. During training, the framework computes a distance measure between the $\hat{\mathbf{y}}$ and the ground truth $\mathbf{y}_\textrm{GT}$ and backpropagates the softloss $\mathcal{L}_{\textrm{soft}}$. During testing, the approach delivers a solution that imitates the expert behavior, while obeying a set of nonlinear constraints.  }
	\label{fig:motivation}
	\vspace{-0.3cm}
\end{figure}

That raises the question:
\textit{Can we combine offline IL methods like BC with the constraint incorporation of optimal control methods?}

Donti et al. \cite{Donti} present a method for incorporating hard constraints into the training of neural networks. The problem is formulated as a nonlinear program, and evaluated with a simple network architecture. 
Our approach extends their previous work to the robotic IL setting.
The nonlinear program is constructed via direct transcription. Our proposed approach, summarized in Fig. \ref{fig:motivation}, leverages two differentiable procedures to account for equality and inequality constraints and is agnostic to the used network architecture. First, the network predicts a sequence of control vectors, which are explicitly completed to a sequence of states w.r.t. the system dynamics represented as equality constraints. Then, a gradient-based correction accounts for inequality constraints while satisfying the equality constraints.  

\textbf{Contributions.} To summarize, the paper makes the following contributions:
(i) It proposes a general Differentiable Constraint Imitation Learning (DCIL) framework for incorporating constraints, which is agnostic to the particular neural network architecture.
(ii) It demonstrates the approach's effectiveness in one mobile robot and one automated driving environment during closed-loop evaluation. The approach outperforms multiple state-of-the art baselines considering a variety of metrics.

\section{Related Work}
The proposed approach is situated within the broader scope integrating constraints into learning-based approaches and IL in the robotics and automated driving literature. 
This section classifies related work into two major categories.

\textbf{Modification of the Training Loss.} The first class of approaches incorporates constraints by modifying the training loss. A simple approach adds the constraints as weighted penalties to the imitation loss.
 \cite{ChaufferNet2019} proposes an application for automated driving. The work shows that additional loss functions penalizing constraint violations improve the closed-loop performance.
\cite{Nandwani2019} modifies the training process with a primal-dual formulation and converts the constrained optimization problem into an alternating min-max optimization with Lagrangian variables. 
\cite{Zeng_2019_CVPR} uses an energy-based formulation. During training, the loss pushes down the energy of positive samples (close to the expert demonstration) and pulls up the energy-values on negative samples, which violate constraints (e.g., colliding trajectories). 
While these methods are more robust to errors in constraint-specifications, they often fail in OOD scenarios as errors made by the learned model still compound over time. That can lead to unexpected behavior like leaving the driving corridor \cite{Dagger}.

\textbf{Projection onto Feasible Sets.}
The second group of approaches projects the neural network's output onto a solution that is compliant with the constraints.
Instead of predicting a future sequence of states, a neural network predicts a sequence of controls \cite{Cui2020}. Unrolling a dynamics model generates a feasible state trajectory consistent with the robot system dynamics. However, the approach does not account for general nonlinear inequality constraints. \cite{Motional2022} presents an inverse reinforcement learning approach. First, a set of safe trajectories is sampled, and learning is only performed on the safe samples. SafetyNet \cite{SafetyNet} trains an IL planner and proposes a sampling-based fallback layer performing sanity checks. \cite{ZhouIROS2021} proposes a similar approach using quadratic optimization. Other works  incorporate quadratic programs \cite{Amos2017} or convex optimization programs \cite{Agraval2019} as an implicit layer into neural network architectures. These approaches constitute the last layer to project the output to a set of feasible solutions. \cite{Brosowky2021} directly modifies the network architecture by encoding convex polytopes.
Sampling, quadratic optimization and convexity severely restrict the solution space. 

Most closely related to our approach is the work of \cite{Donti}. The authors present a hybrid approach, which accounts for nonconvex, nonlinear constraints. Experiments deal with numerical examples with simple network architectures. We extend this work to the real-world-oriented robot IL setting with more complex architectures for high-dimensional feature spaces. Further, we use an explicit completion by unrolling a robot dynamics model.

Just recently, concurrent works propose approaches which also incorporate nonlinear constraints using Signal Temporal Logic \cite{Leung2022} and differentiable control barrier functions \cite{Xiao2022}, which emphasizes the importance of using nonlinearities. In contrast, our approach relies on a differentiable completion, and gradient-based correction procedure, and the training is guided by auxiliary losses. \cite{Leung2022} evaluates on simple toy examples, whereas our analysis considers a more realistic environment. \cite{Xiao2022} evaluates in real-world experiments but only use a circular robot footprint and object representation, whereas this work evaluates using different constraints. Moreover, our approach is able to resolve incorrect constraints that render the problem infeasible.

\section{Problem Formulation}
Assume robots dynamics described by nonlinear, time-invariant differential equations with time $t\in\mathbb{R}$, state $\mathbf{x} \in \mathcal{X}$ and controls $\mathbf{u} \in \mathcal{U}\subset\mathbb{R}^{n_\textrm{u}}$:
\begin{equation}
	\dot{\mathbf{x}}(t)=\mathbf{f}\big(\mathbf{x}(t),\mathbf{u}(t)\big).
	\label{dynamicsmodel}
\end{equation}
The state space size $\mathcal{X}$ of dimension $n_\textrm{x}$ is the union of an arbitrary number of real spaces and non-Euclidean rotation groups $SO(2)$. 
In addition to the low-dimensional state representation $\mathbf{x}$, assume access to a high-dimensional environment representation $\mathbf{e}\in E\subset\mathbb{R}^{n_\textrm{e}}$ (e.g., a birds-eye-view (BEV) image of the scene).  Further, the system is bounded by a set of nonlinear constraints $\mathbb{C}$ (e.g., by control bounds, rules, or safety constraints).

A (sub-)optimal expert, pursuing a policy $\boldsymbol{\pi}_\textrm{exp}$, controls the robot and generates a dataset $\mathcal{D}=\left\lbrace(\mathbf{x}_i, \mathbf{u}_i, \mathbf{e}_i, \mathbb{C}_i   )\right\rbrace^I_{i=0}$ with $I\in\mathbb{N}^+$ samples. A future trajectory of length $H \in \mathbb{N}^+$ containing states and controls belonging to sample $i$ is given by  $\mathbf{y}_\textrm{GT}=\left[\mathbf{x}^\textrm{T}_{i},\mathbf{u}^\textrm{T}_{i}\dots,\mathbf{x}^\textrm{T}_{i+H}, \mathbf{u}^\textrm{T}_{i+H-1}\right]^\textrm{T} $.
During training, the objective is to find the optimal parameters $\boldsymbol{\uptheta}\in\mathbb{R}^{n_\uptheta}$ under a maximum likelihood estimation:
\begin{equation}
	\boldsymbol{\uptheta}^*=\arg\min_{\boldsymbol{\uptheta}}\mathbb{E}\left[\mathbf{d}\big(\mathbf{y}_\textrm{GT},\hat{\mathbf{y}}\big) \right], 
	\label{maximumlikelihoode}
\end{equation} subject to equation (\ref{dynamicsmodel}) and the constraints $\mathbb{C}$. The function $\mathbf{d}$ denotes a distance measure and  $\hat{\mathbf{y}}=\boldsymbol{\pi}_{\boldsymbol{\uptheta}}(\mathbf{x}_i,\mathbf{e}_i)$ is the output of the function $\boldsymbol{\pi}_{\boldsymbol{\uptheta}}$ parameterized by $\mathbf{\boldsymbol{\uptheta}}$. Function $\boldsymbol{\pi}_{\boldsymbol{\uptheta}}$ is described by a neural network $\mathbf{N}_{\boldsymbol{\uptheta}}$ and the completion $\mathbf{f}_\textrm{compl}$ and correction $\mathbf{f}_\textrm{corr}$ procedure.
During inference, given the environment representation, the robot's goal is to predict a sequence of states and controls compliant with the constraints. In the spirit of an model predictive control (MPC) framework, the first control vector is applied or an underlying tracking controller regulates the robot along the reference.

\section{Constrained Imitation Learning Synthesis}
This section introduces the constrained IL framework. We first show how to construct a nonlinear program (NLP) per sample used for training the network. Afterwards, the solution process is detailed. A general description of the approach is visualized in Fig. \ref{fig:motivation}.

\subsection{Nonlinear Program Formulation}

Direct transcription (see for example \cite{2010_John}) transforms the time-continuous formulation of the constraints $\mathbb{C}$ and Equ. (\ref{dynamicsmodel}) into a nonlinear program per sample. We discretize
the time interval of the future with length $H$ with $t_0 \leq t_1 \leq \dots \leq t_k \leq \dots \leq t_H$ and $k=0,1,\dots, H$. We assume a piecewise constant control $\textbf{u}(t):=\textbf{u}_k=\text{constant}$ for $ t\in\left[t_k,t_k+\Delta t \right)$ , where $\Delta t=t_{k+1}-t_{k}$ for $k=0,1,\dots, H-1$ denotes the time interval. The states at grid points $t_k$ are described by $\mathbf{x}(t_k):=\mathbf{x}_k$ for  $k=0,1,\dots, H$. The forward differences
\begin{equation}
	\mathbf{x}_{k+1} = \mathbf{f}(\mathbf{x}_k, \mathbf{u}_k)
	\label{eq:discdynamicsmodel}
\end{equation}
impose a set of equality constraints $\mathbf{h}(\mathbf{x}_{k+1},\mathbf{x}_k, \mathbf{u}_k)=0$. With a slight abuse of notation, we set $\mathbf{x}_0=\mathbf{x}_i$, and at $t_0=t_i$. Note that index $i$ denotes the measured variables in the dataset, whereas index $k$ describes the variables of the constrained optimization problem (\ref{NLP}). Further, inequalities constraints $\mathbf{g}(\mathbf{x}_k, \mathbf{u}_k)\leq0$ are constructed based on $\mathbb{C}_i$ and are only evaluated at the discrete time steps for $\mathbf{x}_k$ and $ \mathbf{u}_k$.

The resulting NLP per sample is given:
\begin{equation}
	\begin{aligned}
		& \min_{\hat{\mathbf{y}}}
		& & 
		\mathbf{d}\big(\mathbf{y}_\textrm{GT},\hat{\mathbf{y}}\big)   \\
		&\text{subject to} 
		\label{NLP}
	\end{aligned}
\end{equation}
\vspace{-0.1cm}
\begin{equation*}
	\begin{aligned}
		&\mathbf{h}(\mathbf{x}_{k+1},\mathbf{x}_k, \mathbf{u}_k)=0, 	&&k=0,1,\dots, H-1 \\ 
		&\mathbf{g}(\mathbf{x}_k, \mathbf{u}_k)\leq0, &&k=0,1,\dots, H-1\\ 
		&\mathbf{g}(\mathbf{x}_N)\leq0, \\ 
	\end{aligned}
\end{equation*} with optimization vector $\hat{\mathbf{y}}=\left[\mathbf{x}^\textrm{T}_{0},\mathbf{u}^\textrm{T}_{0}\dots,\mathbf{x}^\textrm{T}_{H}, \mathbf{u}^\textrm{T}_{H-1}\right]^\textrm{T} $.

Remember that $\hat{\mathbf{y}}$ is a function of the parameters $\boldsymbol{\uptheta}$. Hence, the complete procedure must be differentiable in order to backpropagate the gradients. The next section will describe such an approach using a modified version of \cite{Donti}.

\subsection{Explicit Equality Completion} 
Instead of directly regressing a trajectory of future states, it is a common practice \cite{Cui2020} to output a sequence of control vectors and unroll a differentiable dynamics model. That is similar to the explicit completion procedure described by \cite{Donti}.  To be precise, the neural network $\mathbf{N}_{\boldsymbol{\uptheta}}$ predicts a sequence of control vectors $\mathbf{u}_N$. The sequence of states $\mathbf{x}_N$ is then computed by iteratively  applying Equ. (\ref{eq:discdynamicsmodel}), starting from the measured state $\mathbf{x}_i$, described by  function $\mathbf{x}_N=\mathbf{f}_\textrm{compl}(\mathbf{u}_N, \mathbf{x}_i)$. The concatenation of both vectors results in $
	\bar{\mathbf{y}} = \left[\mathbf{u}^\textrm{T}_{N},\mathbf{x}_N^\textrm{T}\right]^T
   $

\subsection{Inequality Correction}
The completion process accounts for the equality constraints derived from the discretized robots system dynamics. To further consider the inequality constraints, a differentiable gradient-based correction procedure is applied \cite{Donti}. Here, we take gradient steps along the manifold of states and controls satisfying the equality constraints towards a feasible region.

The gradient-based correction, described by function $\mathbf{f}_\textrm{corr}(\bar{\mathbf{y}})$, is initialized by $\bar{\mathbf{y}} = \left[\mathbf{u}^\textrm{T}_{N},\mathbf{x}_N^\textrm{T}\right]^T$. The approach then calculates the gradients of the inequality constraints w.r.t. the sequence of control vectors  $\mathbf{u}_{N}$ and takes $n_\textrm{grad}$ steps along the gradients. With the learning rate $\gamma\in \mathbb{R}^+$ and abbreviating $\mathbf{f}_\textrm{compl}(\cdot)=\mathbf{f}_\textrm{compl}(\mathbf{u}_N$,$\mathbf{x}_i)$ formally the function is given by: 
\begin{equation}\label{align:Gradientenverfahren_DC3}
	\mathbf{f}_\textrm{corr}\left(\left[\begin{array}{c}
		\mathbf{u}_N \\
		\mathbf{f}_\textrm{compl}(\cdot)
	\end{array}\right]\right)=\left[\begin{array}{c}
		\mathbf{u}_N-\gamma \Delta \mathbf{u}_N \\
		\mathbf{f}_\textrm{compl}(\cdot)-\gamma \Delta \mathbf{f}_\textrm{compl}(\cdot)
	\end{array}\right],
\end{equation} with gradients
\begin{equation}
	\Delta \mathbf{u}_N=\nabla_{\mathbf{u}_N}\left\|\operatorname{ReLU}\left(\boldsymbol{\alpha}\odot\mathbf{g}\left(\left[\begin{array}{c}
		\mathbf{u}_N \\
		\mathbf{f}_\textrm{compl}(\cdot)
	\end{array}\right]\right)\right)\right\|_{2}^{2}, 
\label{equ:gradients}
\end{equation} and 
\begin{equation}
	\Delta \mathbf{f}_\textrm{compl}(\cdot)=\frac{\partial \mathbf{f}_\textrm{compl}(\cdot)}{\partial \mathbf{u}_N} \Delta \mathbf{u}_N.
\end{equation}Equ. (\ref{equ:gradients}) calculates the gradients of the inequality constraints $\mathbf{g}$ (depended on $\mathbf{u}_N$ and $\mathbf{x}_N$). $\mathbf{g}$ is weighted by $\boldsymbol{\alpha}\in \mathbb{R}^{\textrm{a}}$, with $\odot$ as the element-wise product. The norm is squared, leading to a quadratic penalty for inequality violations \cite{Roesmann2017}. The ReLU only activates the penalty when the inequality is violated. For instance, the trajectory of an SDV not violating the lane bounds should not be corrected. The solution of the procedure\footnote{While $\mathbf{f}_\textrm{corr}$ respects the equality constraints $\mathbf{h}$, it could lead to violations of them. However, empirically, we found that penalizing $\mathbf{h}$ in Equ. (\ref{sotloss}) led to mean equality violations in the order of $1\mathrm{e}{-5}$, which seems neglectable in our application.} is $\hat{\mathbf{y}}$. 
The intuition is that the network provides a good initialization that, if at all, violates the constraints slightly. Afterward, $\mathbf{f}_\textrm{corr}(\bar{\mathbf{y}})$ corrects those initialization to satisfy all inequality constraints, such as safety constraints, e.g., lane boundaries. That procedure is similar to \cite{sadatiros2019}, which produces an initial trajectory using sampling-based optimization and fine-tunes it with gradient-based optimization. In contrast, our initialization is learned.

\subsection{Training and Inference}
\label{TrainingInference}
As already noticed by \cite{Donti}, the convergence of gradient-based methods is not guaranteed and depends on initialization. However, if initialized closed to an optimum these methods are highly effective. The softloss for training\footnote{The loss in Equ. (\ref{sotloss}) described in the paper of \cite{Donti} squares the norms of  constraint violations. However, the official implementation (https://github.com/locuslab/DC3) uses the same loss as in this work. The authors of \cite{Donti} verified that the mentioned implementation was used to generate the results of their paper. As later discussed in Section \ref{Discussion}, loss (\ref{sotloss}) also produced better results in our experiments.}, 
\begin{equation}
	\label{sotloss}
	\mathcal{L}_\textrm{soft} = \mathbf{d}\big(\mathbf{y}_\textrm{GT},\hat{\mathbf{y}}\big)
	+ \lambda_{g}\left\|\operatorname{ReLU}\left(\boldsymbol{\alpha}\odot\mathbf{g}(\hat{\mathbf{y}})\right)\right\|_{2}
	+ \lambda_{h}\left\|\mathbf{h}(\hat{\mathbf{y}})\right\|_{2},
\end{equation} enables a feasible or at least nearly feasible initial solution, such that the inequality correction
converges during test time. $\lambda_{g}\in\mathbb{R}$ and $\lambda_{h}\in \mathbb{R}$ are weighting factors. Algorithm \ref{alg:DCIL} summarizes the approach.
\begin{minipage}[t]{\columnwidth}
	\vspace{-0.6cm}
	\begin{algorithm}[H]
		\caption{Deep Constraint Imitation Learning}
		\label{alg:DCIL}
		\begin{algorithmic}[1]
			\Procedure{DCIL}{$\mathbf{e}_i,\mathbf{x}_i}$ 
			\State \textbf{compute} initial sequence of controls  $\mathbf{u}_N = \mathbf{N}_{\boldsymbol{\uptheta}}(\mathbf{e}_i) $
			\State \textbf{complete} to  $\bar{\mathbf{y}} =\left[\mathbf{u}^\textrm{T}_{N},\mathbf{x}_N^\textrm{T}\right]^T $ with $\mathbf{f}_\textrm{compl}(\mathbf{u}_N, \mathbf{x}_i)$  
			\State \textbf{correct} to estimated solution $\hat{\mathbf{y}} =
			 \mathbf{f}_\textrm{corr}(\bar{\mathbf{y}})$ (function applied $n_\textrm{grad}$ times)
			\If{train}
				\State \textbf{compute} loss (\ref{sotloss}) and \textbf{update} $\theta$
			\Else
				\State \textbf{return} $\hat{\mathbf{y}}$
			\EndIf

			\EndProcedure
		\end{algorithmic}
	\end{algorithm}
\end{minipage}

\begin{figure*}
	\centering
	\begin{subfigure}[b]{0.24\textwidth}
		\centering
		
		\includegraphics[width=\textwidth,trim={2cm 2cm 0 0},clip]{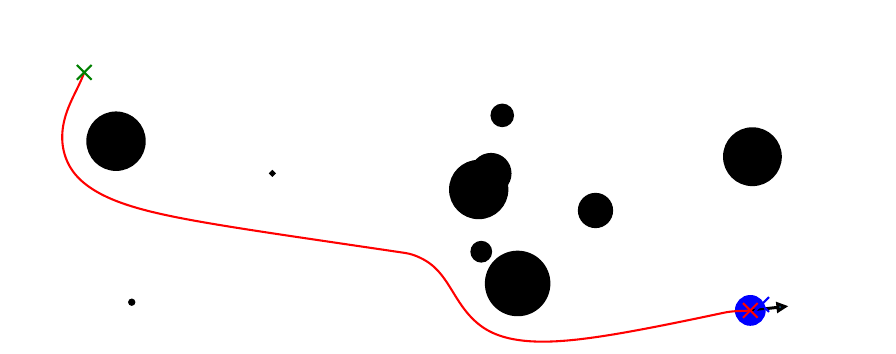}	
		\caption{}
		\label{MRE}
	\end{subfigure}
	\begin{subfigure}[b]{0.15\textwidth}
		\centering
		\includegraphics[width=\textwidth]{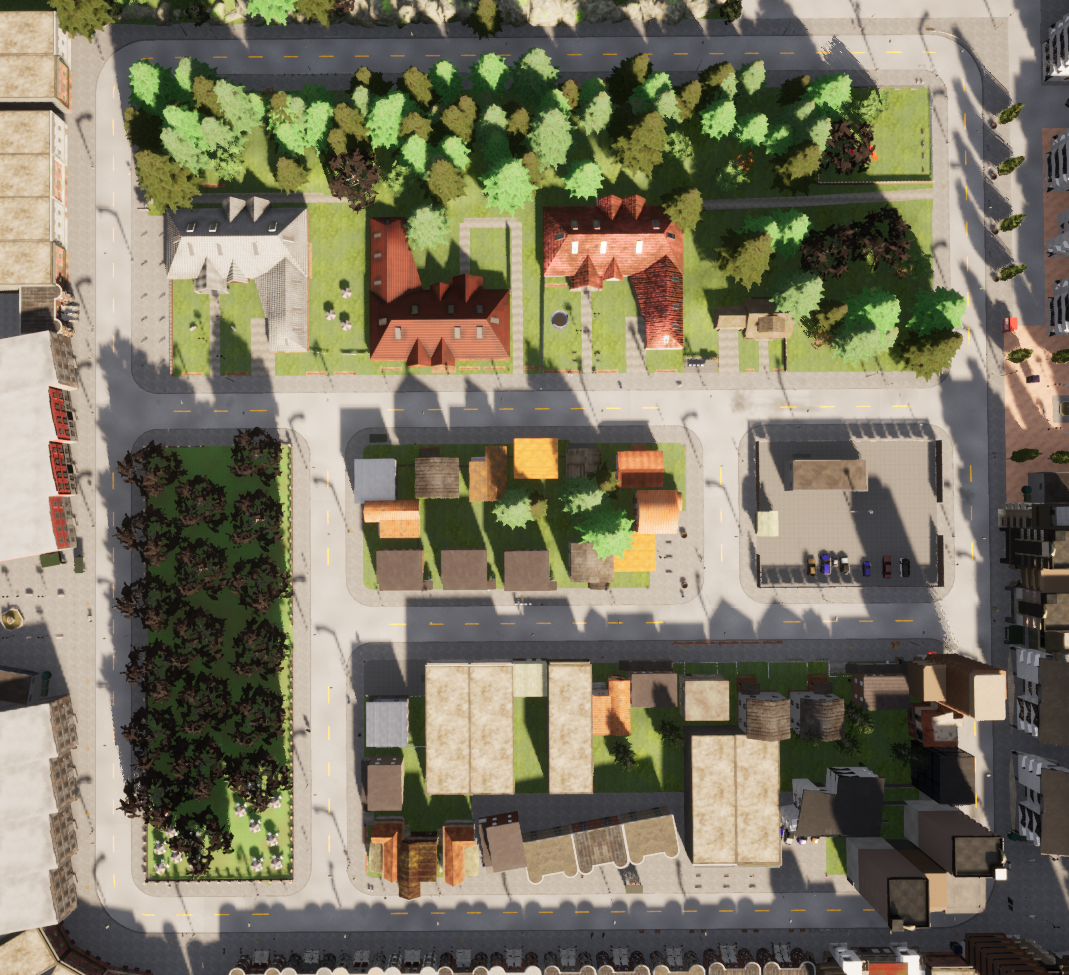}
		\caption{}
		\label{CARLA}
	\end{subfigure}
	\begin{subfigure}[b]{0.55\textwidth}
		\centering
		\includegraphics[width=\textwidth]{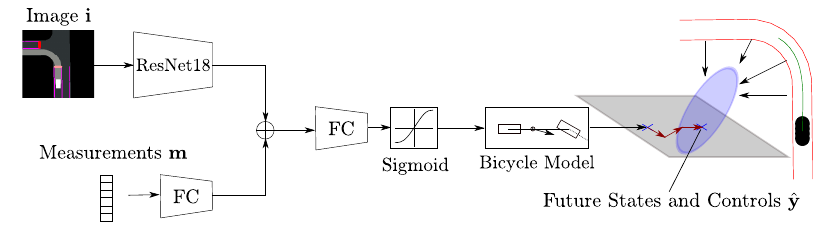}
		\caption{}
		\label{fig::NetworkArchitecture}
	\end{subfigure}
	\label{environments}
	\vspace{-0.2cm}
	\caption{(a) Mobile robot environment. Red visualizes a demonstration trajectory navigating from the green start point to the blue goal region, avoiding random obstacles (black). (b) Self-driving environment. (c) Network architecture in the SDE.}
	\vspace{-0.5cm}
\end{figure*}

\section{Experimental Evaluation}
This section evaluates the proposed approach in one mobile robot and one automated driving environment. This section addresses the following research questions:
\textit{Q1}: Does the approach improve the closed-loop performance of IL methods?
\textit{Q2}: How does the approach deal with incorrect constrain specifications?

\textbf{Environments.} The environments used for evaluation are visualized in Fig. \ref{MRE}) and b). 

\textit{Mobile Robot Environment} (MRE): In the first environment data is collected by controlling a mobile robot with radius $r_\textrm{robot}=1\si{\metre}$ using the Dynamic Window Approach \cite{DWA}. The task during demonstrations is to navigate from a random start to a random goal location in the shortest time possible while avoiding collisions with circular shaped obstacles. Objects are randomly located with varying radii $r_\textrm{c} \in\left[ 0.1, 3\right]\si{\metre} $. The dataset contains $838$ episodes ($69638$ samples), which were spitted in $83.3\%$ training, and $8.3\%$ validation and test each samples. This work evaluates closed-loop on another $76$ \textit{unseen} test episodes.

\textit{Self-Driving Environment} (SDE): The second environment uses CARLA, a realistic automated driving simulator \cite{CARLA}. The CARLA Roach agent \cite{Roach} collects training data. Further additive noise is applied to make the demonstrations more diverse, but sub-optimal. The dataset contains $120$ episodes ($174275$ samples) from Town01, using the same ratios as in the MRE. We test on $25$ random routes from a \textit{different} environment (Town02) using the scenarios of the CARLA NoCrash-Challenge (Empty) \cite{Codevilla2019} following the standardized evaluation protocol of the CALRA leaderboard. The SDV's task it to follow the routes while avoiding collisions and obeying traffic lights.

\textbf{Baselines.} This work benchmarks against the following baselines. \textit{IL}: The traditional imitation learning  directly regresses a future state trajectory.  \textit{DKM}: An IL approach \cite{Cui2020} predicting a sequence of control vectors bounded by a sigmoid layer. The controls and a dynamics model are then used to unroll a future state trajectory. \textit{DKM$\leq$}: DKM with an additional gradient-based correction procedure only applied during test time, similar to a fallback layer as in \cite{SafetyNet}. \textit{SL}: The same approach as IL trained using the softloss $\mathcal{L}_\textrm{soft}$. That is similar to \cite{ChaufferNet2019}, but here the soft constraints are not computed in image space.

\textbf{Metrics.} 
The MRE uses the following metrics: \textit{Goal Reaching Rate} (GRR): Rate of reached goals. \textit{Collision Rate} (CR): Rate of collision-prone episodes. \textit{Time}: Percentage of the agents completion time relative to the expert trajectory. This metric measures efficiency. \textit{Kinematic Constraint Violations} (KCV): Summed number of constraint violations (tolerance: $1\mathrm{e}{-4}$) of velocity, angular velocity, acceleration, and angular acceleration.

The SDE uses the metrics of the official CARLA Leaderboard Benchmark as described in \cite{Roach}. We focus on closed-loop metrics as open-loop metrics can be a poor indicator to the actual task performance of robot policies \cite{Xiao2022}.

\subsection{Implementation}
The method is agnostic and not restricted to the specific design choices made here.

\textbf{State and Controls.}
A robot-centric coordinate system describes the state.
\textit{MRE}: The state $\mathbf{x}$ is defined by a 2-D position with $x,y \in \mathbb{R}$ and $\phi \in SO(2)$. The robot controls $\mathbf{u}$ are described by a forward $v\in\mathbb{R}$ and rotational velocity $\omega\in\mathbb{R}$.
During testing, the flatness property of the unicycle model is used to compute the control values based on the predicted state trajectory. One could also directly use the predicted control values in a MPC formulation. However, the IL baseline only predicts a state sequence. Therefore, for a fair comparison, DCIL uses the same control strategy.

\textit{SDE}: State  $\mathbf{x}$ and  controls $\mathbf{u}$ are the same as in \cite{Cui2020}.
During the evaluation, two PID controllers track the predicted state sequence of all methods.

\textbf{Inputs.}
In both environments the input of the neural network $\mathbf{e}_i$ constitutes of an image $\mathbf{i}\in \mathbb{R}^{a\times b \times c}$ with resolution $\textrm{res}\in\mathbb{R}$ and a measurement vector $\mathbf{m}\in \mathbb{R}^{n_\textrm{m}}$, which is a common representation in automated driving applications \cite{Cui2020, RALDiehl}. 

\textit{MRE:} The robot centric image has dimensions $a = b = 128\,\textrm{px}$ with $\textrm{res}=10\,\frac{\textrm{px}}{\si{\metre}}$. One channel $c=1$ describes the binary occupancy information. $\mathbf{m}$ contains the current $v$ and $\omega$. It is further described by the distance $d_\textrm{goal}\in\mathbb{R}$ and heading $\theta_\textrm{goal} \in SO(2)$ w.r.t. the goal point. The dimension of the estimated control and state sequence $\hat{\mathbf{y}}$, is defined by $H=10$ with time interval $\Delta t=0.3\, \si{\second}$. 

\textit{SDE:}: The image has dimensions $a = b = 192\,\textrm{px}$ with $\textrm{res}=5\,\frac{\textrm{px}}{\si{\metre}}$. The SDV is centered in all images at $40\,\textrm{px}$ above the bottom. Different semantic classes from the work of \cite{Roach} are color coded using the RGB channels with $c=3$ as visualized in \ref{fig::NetworkArchitecture}). $\mathbf{m}$ contains the current 2-D velocity $\mathbf{v}\in\mathbb{R}^2$, acceleration $\mathbf{a}\in\mathbb{R}^2$, and current speed limit $v_\textrm{max}$. We set $H=20$ and $\Delta t=0.2\,\si{\second}$. 

\textbf{Constraints.}
\textit{MRE:} For the dynamics, which constitute the equality constraints, we use a unicycle model as in \cite{ChungLee20021}. This work applies box constraints such that $v\in [-0.5,1]\frac{\textrm{m}}{\textrm{s}}$, $\omega\in  [-0.70,0.70]\frac{\textrm{rad}}{\textrm{s}}$, $a\in [-0.2,0.2]\frac{\textrm{m}}{\textrm{s}^2}$,$\dot{\omega}\in  [-0.70,0.70]\frac{\textrm{rad}}{\textrm{s}^2}$.
 The acceleration $a$ and angular acceleration $\dot{\omega}$ are computed from finite differences. For collision avoidance, we compute euclidean distances $d_\textrm{obst}$ between the robot and the obstacles, as both footprints are circles. Then the state of every predicted time step is constrained by $d_\textrm{obst}>r_\textrm{robot}+r_\textrm{c}+0.1\,\si{\metre}$.  As the algorithm requires a fixed number of constraints, it uses the three closest obstacles in the front half level of the robot.

\textit{SDE}: In the CARLA experiments, this work uses the extended bicycle  model for the dynamics (\ref{eq:discdynamicsmodel}) as \cite{Cui2020} with vehicle sizes of a Lincoln MKZ. We bound the velocity by $v_\textrm{max}=\SI{8.33}{\frac{m}{s}}$. Further the control accelerations are constraint by  $a\in [-8,4]\frac{\textrm{m}}{\textrm{s}^2}$, and the control steering angles by $\delta \in [-1,1]\,\textrm{rad}$.
For collision avoidance, this work constructs a polyline-based driving corridor as in \cite{Ziegler} using the high-level route. Four circles approximate the vehicle footprint. At every gradient step, the algorithm estimates the shortest distance to the left and right lane boundary for every predicted time step $k$ and every circle.  Further logical constraints for traffic lights are imposed. If a traffic light is yellow or red, it constructs a stop line in front of the vehicle. Otherwise, due to the required fixed number of constraints, this line is created far away not affecting the correction step. For the stop line and the driving corridor, the inequality constraints are described as point-line distances.
For a visual example of the constraints refer to Fig. \ref{fig::NetworkArchitecture}).

\textbf{Loss.}
\textit{MRE}: Let subscript $\hat{}$ define the estimated state and control of $\mathbf{y}$. Based on related work \cite{Cui2020}, losses are:  
\begin{equation}
	\begin{aligned}
 	\mathbf{d}\big(\mathbf{y}_\textrm{GT},\hat{\mathbf{y}}\big)=
 	\sum_{k=1}^{H}(\hat{x}_{k}-x_{k,\textrm{GT}})^2+(\hat{y}_{k}-y_{k,\textrm{GT}})^2\\
 	+\left(\cos(\hat{\phi}_{k})-\cos(\phi_{k,\textrm{GT}})\right)^2+\left(\sin(\hat{\phi}_{k})-\sin(\phi_{k,\textrm{GT}})\right)^2.
	\end{aligned}
\label{loss}
\end{equation}
\textit{SDE}:
 \cite{Janjos2021} showed that it its also beneficial to use an regularization term in the form of a inverse dynamics model. We follow this approach by adding the term to Equ. (\ref{loss}).

\textbf{Network Architecture and Parameters.}
\textit{MRE}:
The binary image $\mathbf{i}$ is encoded by a LeNet \cite{LeNet} outputting a latent vector, which is concatenated with the measurement vector $\mathbf{m}$. Afterwards, the result is processed by the same 2-layer fully connected network (FCN) of \cite{Donti}. The predicted control sequence is bounded by a sigmoid layer and passed to the completion and correction step. In both environments we choose the hyperparameters by grid searches. In MRE we choose: $\lambda_g = \lambda_h= 0.5$, $\boldsymbol{\alpha}=\mathds{1}$.

\textit{SDE}:
A ResNet18 \cite{Resnet} first encodes the RGB image $\mathbf{i}$ and the measurement vector $\mathbf{m}$ is encoded by a 1-layer FCN . The concatenation of both encodings is passed to the previously described 2-layer FCN. Fig. \ref{fig::NetworkArchitecture}) visualizes the network architecture.  
In SDE we choose: $\lambda_g = 5 $, $\lambda_h= 5 $, $\lambda_u= 1 $. Vector $\boldsymbol{\alpha}$ is defined by  weighting factors for the different inequality constraints. Collision is weighted by factor $\alpha_\textrm{c}=1$, stopping line violations by $\alpha_\textrm{s}=2$ and bounds on kinematic values by $\alpha_\textrm{k}=1$.
Both experiments use $\gamma=1\textrm{e}-3$ and $n_\textrm{grad}=5$.
For a fair comparison, we ran grid searches for all baselines. 

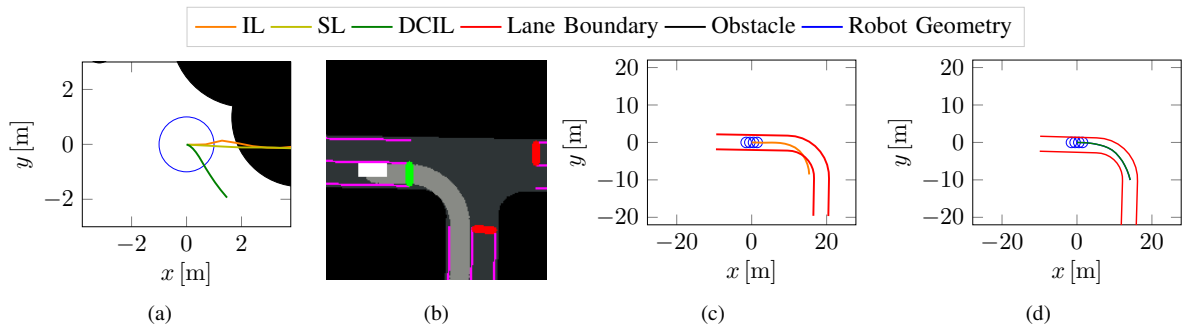
\begin{figure*}[t]
	
	\centering
	\resizebox{0.9\textwidth}{!}{
	\begin{subfigure}[b]{0.26\textwidth}
		\centering
		
\begin{tikzpicture}

\definecolor{darkgray176}{RGB}{176,176,176}
\definecolor{darkviolet1910191}{RGB}{191,0,191}
\definecolor{goldenrod1911910}{RGB}{191,191,0}
\definecolor{green01270}{RGB}{0,127,0}
\definecolor{lightgray204}{RGB}{204,204,204}
\definecolor{darkturquoise0191191}{RGB}{0,191,191}

\begin{axis}[
	width=\textwidth,
	xlabel shift = -3 pt,
	ylabel shift = -3 pt,
legend cell align={left},
legend style={
	fill opacity=0.8,
	draw opacity=1,
	text opacity=1,
	at={(0.5,1.335)},
	anchor=north west,
	draw=lightgray204,
	legend columns=-1,
	axis equal
}, 
x grid style={darkgray176},
xlabel={$x \left[ \si{\metre}\right] $ },
xmin=-3, xmax=3,
y grid style={darkgray176},
ylabel={$y \left[ \si{\metre} \right] $ },
ymin=-3, ymax=3,
]
\draw[draw=blue] (axis cs:0,0) circle (1);
\draw[draw=black,fill=black] (axis cs:4.17253252105903,0.981603236298873) circle (2.52409137437313);
\draw[draw=black,fill=black] (axis cs:17.4106932531882,11.6093649998003) circle (1.52026860696686);
\draw[draw=black,fill=black] (axis cs:2.20117714567247,4.12148896483963) circle (2.77075375707685);
\draw[draw=black,fill=black] (axis cs:3.20953750029467,7.81152003764515) circle (1.35997070921599);
\draw[draw=black,fill=black] (axis cs:-3.17159747442486,3.40939878602749) circle (0.444554377590418);
\draw[draw=black,fill=black] (axis cs:-5.48418877842259,-16.9137602209799) circle (0.95756219299926);
\draw[draw=black,fill=black] (axis cs:2.19689704843973,-11.7370712592342) circle (2.2188598816561);

\addplot [thick, orange]
table {%
0.0816887653900225 -0.0164213328475397
0.692712645632406 -0.00201496700269639
1.29147342514477 0.135911833439275
1.82945919891502 0.0500174696638618
2.40928992974395 -0.0651965521908238
2.93338131935403 -0.11875869299501
3.39167085731809 -0.128158860930286
3.91432333051642 -0.0827423604960992
4.44774279905006 -0.0991381457269958
4.95196513996346 -0.0978104826840697
5.36062065111293 -0.0297290897642301
};
\addlegendentry{IL}
\addplot [thick, goldenrod1911910]
table {%
0.0412639482444254 -0.00912445530953448
0.415619983632285 -0.0307771436992457
0.808442347481241 -0.049846070773335
1.22157877437475 -0.0747019596857913
1.61841537726955 -0.0949169348052482
2.00841464644098 -0.112154210021867
2.40566878804318 -0.112634395523484
2.78712496469988 -0.113851380174747
3.1749734141113 -0.119348189693964
3.55121064292291 -0.134017751350984
3.92647165336044 -0.143115246249216
};
\addlegendentry{SL}

\addplot [thick, green01270]
table {%
	0 0
	0.113331575142989 -0.0531589248488295
	0.229245978429954 -0.152169739603211
	0.342041002692298 -0.28656895899754
	0.455868313967822 -0.452776242689459
	0.574127915610683 -0.653955842136308
	0.70615003207867 -0.881120372146972
	0.86060248150844 -1.12358940427895
	1.03727299959008 -1.38676505892134
	1.24126581525907 -1.6590305936289
	1.47358413995262 -1.93878083228993
};

\addlegendentry{DCIL}
\addplot [thick, red]
table {%
10 10 
11 11
};
\addlegendentry{Lane Boundary}

\addplot [thick, black]
table {
	10 10 
	11 11
};
\addlegendentry{Obstacle}

\addplot [thick, blue]
table {
	10 10 
	11 11
};
\addlegendentry{Robot Geometry}

\end{axis}

\end{tikzpicture}	
		\vspace{-0.5cm}
		\caption{}
		\label{MREQual}

	\end{subfigure}
	\begin{subfigure}[b]{0.18\textwidth}
				\centering
		
		\includegraphics[width=\textwidth, angle=-90]{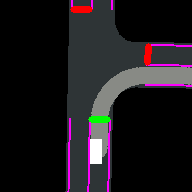}
		\vspace{0.13cm}
		\caption{}
		\label{QualCompBEV}
		
	\end{subfigure}
	\hfill
	\begin{subfigure}[b]{0.26\textwidth}
		
		\centering
\begin{tikzpicture}

\definecolor{darkgray176}{RGB}{176,176,176}
\definecolor{goldenrod1911910}{RGB}{191,191,0}
\definecolor{lightgray204}{RGB}{204,204,204}
\definecolor{darkturquoise0191191}{RGB}{0,191,191}

\begin{axis}[
	width=\textwidth,
xlabel shift = -3 pt,
ylabel shift = -3 pt,
legend cell align={left},
legend style={
	fill opacity=0.8,
	draw opacity=1,
	text opacity=1,
	at={(0.5,1.335)},
	anchor=north west,
	draw=lightgray204,
	legend columns=-1,
}, 
x grid style={darkgray176},
xlabel={$x \left[ \si{\metre}\right] $ },
xmin=-22, xmax=22,
y grid style={darkgray176},
ylabel={$y \left[ \si{\metre} \right] $ },
ymin=-22, ymax=22,
axis equal
]
\draw[draw=blue] (axis cs:-1.47050499916077,0) circle (1.4);
\draw[draw=blue] (axis cs:-0.490168333053589,0) circle (1.4);
\draw[draw=blue] (axis cs:0.490168333053589,0) circle (1.4);
\draw[draw=blue] (axis cs:1.47050499916077,0) circle (1.4);
\addplot [thick, orange]
table {%
0 0
0.414751471607666 -0.0181370193563197
0.904334074090058 -0.0150667423017588
1.43986917541389 -0.0277532170332817
2.04336703518901 -0.0455792646847927
2.74564570075538 -0.0686041796735514
3.54274046537796 -0.0721536146600759
4.40894928701411 -0.0571096663459549
5.36531991230852 -0.0423872803492743
6.42308980813091 -0.0687675152242526
7.60173490172254 -0.144215904259752
8.88821673933548 -0.349406094225419
10.2396290739096 -0.743604774664651
11.5850958399336 -1.36709016668273
12.7650878002345 -2.28998668245677
13.7019145124668 -3.40534917851267
14.3744892433529 -4.62047963215971
14.8855851021532 -5.88754925646593
15.2230638286831 -7.21500742237283
15.347639514793 -8.55120360854824
};
\addplot [thick, red]
table {%
-9.45892715454102 2.16353583335876
-7.45925760269165 2.12757515907288
-5.45958089828491 2.09161448478699
-3.45990443229675 2.05565357208252
-1.46022796630859 2.01969265937805
0.5394486784935 1.98373186588287
2.5391252040863 1.94777095317841
4.53880167007446 1.91181015968323
6.5384783744812 1.87584936618805
8.53816509246826 1.83961355686188
9.63797760009766 1.81937289237976
11.6267309188843 1.63419985771179
13.549168586731 1.09276390075684
15.3413639068604 0.211346954107285
16.9437713623047 -0.980783879756927
18.3039398193359 -2.44337797164917
19.3810119628906 -4.12537336349487
20.1409111022949 -5.97250270843506
20.5592555999756 -7.92554330825806
20.6361541748047 -9.6681022644043
20.6007137298584 -11.6677875518799
20.5652713775635 -13.6674737930298
20.529821395874 -15.6671600341797
20.4943790435791 -17.6668453216553
20.4589366912842 -19.6665325164795
};
\addplot [thick, red, forget plot]
table {%
-9.53084850311279 -1.83581745624542
-7.53117942810059 -1.87177813053131
-5.53150272369385 -1.90773904323578
-3.53182625770569 -1.94369983673096
-1.53214967250824 -1.97966063022614
0.467526882886887 -2.01562142372131
2.46720337867737 -2.05158233642578
4.46688032150269 -2.08754324913025
6.46655654907227 -2.12350416183472
8.46451950073242 -2.15970897674561
9.56433200836182 -2.17994976043701
10.8978233337402 -2.29883480072021
12.1183757781982 -2.64257431030273
13.2562303543091 -3.20218586921692
14.2735805511475 -3.95904636383057
15.1420001983643 -4.89329671859741
15.832893371582 -5.97222375869751
16.3203468322754 -7.15709018707275
16.5886917114258 -8.40988922119141
16.6367835998535 -9.59721755981445
16.6013412475586 -11.5969038009644
16.5658988952637 -13.5965900421143
16.5304489135742 -15.5962753295898
16.4950065612793 -17.5959606170654
16.4595642089844 -19.5956478118896
};
\end{axis}

\end{tikzpicture}
		\vspace{-0.05cm}
		\caption{}
		\label{QualCompIL}

	\end{subfigure}
	\begin{subfigure}[b]{0.26\textwidth}
		\centering
\begin{tikzpicture}

\definecolor{darkgray176}{RGB}{176,176,176}
\definecolor{goldenrod1911910}{RGB}{191,191,0}
\definecolor{green01270}{RGB}{0,127,0}
\definecolor{lightgray204}{RGB}{204,204,204}

\begin{axis}[
width=\textwidth,
xlabel shift = -3 pt,
ylabel shift = -3 pt,
legend cell align={left},
legend style={
	fill opacity=0.8,
	draw opacity=1,
	text opacity=1,
	at={(0.5,1.335)},
	anchor=north west,
	draw=lightgray204,
	legend columns=-1,
	axis equal
}, 
x grid style={darkgray176},
xlabel={$x \left[ \si{\metre}\right] $ },
xmin=-22, xmax=22,
y grid style={darkgray176},
ylabel={$y \left[ \si{\metre} \right] $ },
ymin=-22, ymax=22,
axis equal
]
\draw[draw=blue] (axis cs:-1.47050499916077,0) circle (1.4);
\draw[draw=blue] (axis cs:-0.490168333053589,0) circle (1.4);
\draw[draw=blue] (axis cs:0.490168333053589,0) circle (1.4);
\draw[draw=blue] (axis cs:1.47050499916077,0) circle (1.4);
\addplot [semithick, blue]
table {%
0 0
0.432823756388139 -0.0188077190901869
0.889131272977227 -0.0348620507098069
1.40862888178893 -0.042901601281855
1.99815714390475 -0.0729965928754061
2.6589470464684 -0.105202193133236
3.40349180028455 -0.190824394702884
4.21816828686314 -0.307973892095656
5.1056940311253 -0.466659767856126
6.09550502542913 -0.695320599472554
7.18017121098416 -0.995705588292891
8.29510329106023 -1.44469725138016
9.38934581486026 -2.06015269355869
10.4008495913463 -2.89087293342253
11.277870320214 -3.90791100691801
12.0936948837491 -4.99384495930624
12.7546418443794 -6.18225630426534
13.3419310547143 -7.42134556285557
13.820592917863 -8.71966564324194
14.1890294774565 -10.0593371445515
};

\addplot [semithick, green01270]
table {%
0 0
0.432827473806111 -0.0187220597421457
0.889144834467086 -0.0346328369856976
1.40865519601654 -0.0424346477848199
1.99821564642311 -0.0721497056070614
2.65904883746656 -0.103795358941807
3.40370572661223 -0.188605469282219
4.21856089053424 -0.304658941753122
5.10636814262736 -0.461898976973547
6.09665296161756 -0.688614850912188
7.18204510580471 -0.986470645200318
8.29822780063206 -1.43242318975576
9.39440630663606 -2.04448690868044
10.4088008314669 -2.87172556298922
11.2896131403292 -3.88552568268274
12.1096094785061 -4.9683554852995
12.7751981334472 -6.15421275464968
13.3673215782455 -7.39103737719966
13.8510332990251 -8.68752135676576
14.2246707699701 -10.0257874752129
};

\addplot [semithick, red]
table {%
-9.82388687133789 1.64787888526917
-7.82468509674072 1.59138417243958
-5.82548332214355 1.5348893404007
-3.82628130912781 1.47839462757111
-1.82707941532135 1.42189991474152
0.172122478485107 1.36540508270264
2.17132449150085 1.30891036987305
4.17053365707397 1.25214064121246
5.27008056640625 1.22060632705688
7.25682687759399 1.01501953601837
9.17360305786133 0.453869551420212
10.9566516876221 -0.445906013250351
12.5467319488525 -1.65443003177643
13.8918085098267 -3.13091516494751
14.9515514373779 -4.82388305664062
15.6924419403076 -6.67871856689453
16.0907077789307 -8.63595294952393
16.1497077941895 -10.3792085647583
16.0937309265137 -12.3784255981445
16.0377559661865 -14.3776416778564
15.9817714691162 -16.3768577575684
15.9257955551147 -18.3760738372803
15.8698196411133 -20.3752918243408
15.8138437271118 -22.3745079040527
15.757866859436 -24.3737239837646
};

\addplot [semithick, red, forget plot]
table {%
-9.93687629699707 -2.35052490234375
-7.9376745223999 -2.40701961517334
-5.93847274780273 -2.46351432800293
-3.93927073478699 -2.52000904083252
-1.94006884098053 -2.57650399208069
0.0591331161558628 -2.63299870491028
2.05833506584167 -2.68949341773987
4.05582094192505 -2.7462146282196
5.15536737442017 -2.77774882316589
6.48756790161133 -2.91032195091248
7.7045259475708 -3.26657795906067
8.83657264709473 -3.83784532546997
9.84609699249268 -4.60511350631714
10.7048768997192 -5.5482325553894
11.384654045105 -6.63419818878174
11.8599128723145 -7.82400798797607
12.1153793334961 -9.07949733734131
12.1512746810913 -10.2672567367554
12.0952987670898 -12.2664728164673
12.0393218994141 -14.2656898498535
11.9833383560181 -16.2649059295654
11.9273624420166 -18.2641220092773
11.8713865280151 -20.2633380889893
11.8154106140137 -22.2625560760498
11.7594337463379 -24.2617721557617
};

\end{axis}

\end{tikzpicture} 
		\vspace{-0.05cm}
		\caption{}
		\label{QualCompDCIL}
		
	\end{subfigure}}
	\caption{Qualitative comparison. (a) Open-loop prediction results of the different methods in the MRE. (b) BEV input image of the neural network representing in CARLA. White pixels denote the SDV, black static obstacles, dark grey the road, light grey the route, and violet lane markings. Traffic lights are visualized by green or red color. The image is rotated by 90 degree (c) Constraint plot of the IL agent during closed-loop control. (d) DCIL during closed-loop control. }
	\vspace{-0.4cm}	
	
\end{figure*}

\begin{table}[t]
	\caption{Closed-loop performance of all methods using unseen test scenarios the in mobile robot environment. }
	\centering
	\begin{tabular}{lcccc}
		\toprule
		Methode & GRR  & CR  & Time & KCV   \\ 
		\cmidrule(r){2-5}
  
		& [$\%$], $\uparrow$ & [$\%$], $\downarrow$  & [$\%$], $\downarrow$ & [$\%$ ($\#$)], $\downarrow$  \\ 
		\midrule

				IL  & $\textbf{100}$ & $3.94$ & $106$ & $7.20\,(4600)$  \\ 
				SL  & $92$ & $6.57$ & $117$ & $2.06\, (1317)$  \\ 
				\midrule
				DCIL &  $\textbf{100}$ & $\textbf{0.00}$ & $\textbf{105}$ & $\textbf{0.12}\,(\textbf{89})$ \\ 
		
		\bottomrule
	\end{tabular}
	\label{tbl:MRE}
	\vspace{-0.5cm}
\end{table}

\begin{table*}
	\caption{Closed-loop performance of all methods using unseen test routes in Town02 from CARLA-NoCrash. $\downarrow$ indicates a lower number is better and $\uparrow$ vice versa. Bold numbers indicate the best results.}
	\centering
	\begin{tabular}{lccccccccc}
		\toprule
		& Sucess  & Driving  & Route & Infraction  & Collision  & Red light  & Agent & Outside & Wrong \\ 
		Methode    & rate & score & completion & penalty & layout & infraction & blocked & of lane & lane \\
		\cmidrule(r){2-5}
		\cmidrule(r){6-10}   
		& [$\%$], $\uparrow$ & [$\%$], $\uparrow$  & [$\%$], $\uparrow$ & [$\%$], $\uparrow$ & [\#/Km], $\downarrow$  & [\#/Km], $\downarrow$  & [\#/Km], $\downarrow$ & [\#/Km], $\downarrow$ & [\#/Km], $\downarrow$ \\ 
		\midrule
		IL  & $36$ & $50.18$ & $44$  & $95.65$ & $1.44$ & $0.42$ & $525.00$ & $0.06$ & $5.73$ \\ 
		IL++  & $52$ & $62.33$ & $80$  & $98.84$ & $5.70$ & $0.27$ & $94.28$ & $3.39$ & $9.41$ \\ 
		DKM  & $76$ & $76.69$ & $92$  & $98.94$ & $1.40$ & $0.11$ & $6.90$ & $\textbf{0.00}$ & $8.85$ \\ 
		DKM $\leq$   &$76$ & $87{,}66$ & $96$  & $98.94$ & $0.94$ & $0.13$ & $3.26$ & $0.15$ & $0.09$ \\ 
 
		SL  & $92$ & $96.09$ & $\textbf{100}$  & $\textbf{100.00}$ & $0.35$ & $\textbf{0.00}$ & $\textbf{0.00}$ & $\textbf{0.00}$ & $0.21$ \\ 
		\midrule
		DCIL &  $\textbf{96}$ & $\textbf{97.40}$ & $\textbf{100}$  & $98.94$ & $\textbf{0.22}$ & $0.18$ & $\textbf{0.00}$ & $\textbf{0.00}$ & $\textbf{0.00}$ \\ 
		\bottomrule
	\end{tabular}
	\label{tbl:SDE}
	\vspace{-0.4cm}
\end{table*}

\subsection{MRE Results}
To answer \textit{Q1}, DCIL is compared against the described baselines. Table \ref{tbl:MRE} visualizes the results. DCIL outperforms all baselines in all metrics and it is the only one, which reaches all goals and without any collisions. Moreover, compared to the normal IL baseline, the number of kinematic constraint violations is reduced by a factor of $51.69$. A qualitative comparison in an exemplary scenario is visualized in Fig. \ref{MREQual}). DCIL is the only method planning a collision-free trajectory. This can be attributed to the correction procedure acting as a safety layer.

\subsection{SDE Results}
Again considering question \textit{Q1}, refer to the quantitative comparison of Table \ref{tbl:SDE}. Note that IL++ describes the same approach as IL, but uses a PID controller which takes more time to tune, such that the evaluation favors IL++.  However, DCIL performs best in the different metrics of the CARLA leaderboard. We observed that the other baselines (IL, DKM, SL) often plan trajectories, that divert  from the route or onto the opposite. That is explained by their behavior under distribution shifts. Fig. \ref{QualCompIL}) illustrates such an qualitative result using the IL method. The closed-loop metrics\footnote{The closed-loop performance also depends on the underlying tracking controller. Even if the planned trajectory obeys all constraints, an inadequate PID controller could lead to lane boundary violations or red light infractions. For instance, DCIL violates one red traffic light.} in Table \ref{tbl:SDE} also underline the described failures of the baselines. 

\begin{figure}

	\centering
	\begin{subfigure}[b]{0.75\columnwidth}
		\centering
		
		\includegraphics[width=\textwidth]{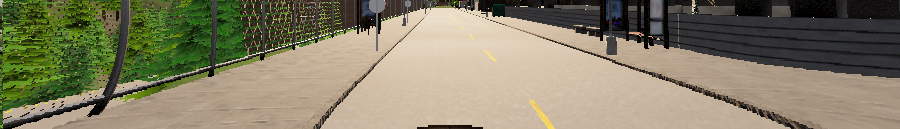}
		\caption{}
		\label{Camera image}
	\end{subfigure}

	\begin{subfigure}[b]{0.22\textwidth}
		\centering
		
\begin{tikzpicture}

\definecolor{darkgray176}{RGB}{176,176,176}
\definecolor{goldenrod1911910}{RGB}{191,191,0}
\definecolor{green01270}{RGB}{0,127,0}
\definecolor{lightgray204}{RGB}{204,204,204}

\begin{axis}[
	width=\textwidth,
xlabel shift = -3 pt,
ylabel shift = -3 pt,
legend cell align={left},
legend style={
	fill opacity=0.8,
	draw opacity=1,
	text opacity=1,
	at={(0.5,1.335)},
	anchor=north west,
	draw=lightgray204,
	legend columns=-1,
	axis equal
}, 
x grid style={darkgray176},
xlabel={$x \left[ \si{\metre}\right] $ },
xmin=-22, xmax=22,
y grid style={darkgray176},
ylabel={$y \left[ \si{\metre} \right] $ },
ymin=-22, ymax=22,
]
\draw[draw=blue] (axis cs:-1.47050499916077,0) circle (1.4);
\draw[draw=blue] (axis cs:-0.490168333053589,0) circle (1.4);
\draw[draw=blue] (axis cs:0.490168333053589,0) circle (1.4);
\draw[draw=blue] (axis cs:1.47050499916077,0) circle (1.4);

\addplot [semithick, green01270]
table {%
0 0
0.855046453612956 -0.0852441913882191
1.84588886470255 -0.239645615645484
2.93970791779445 -0.415872764553172
4.10532370046829 -0.587574216084551
5.34063549320896 -0.765232586411732
6.63028332415263 -0.983616892349154
7.99180025556718 -1.21109168308069
9.43037273622913 -1.34840294402488
10.9255345352809 -1.46224791241036
12.4528175920941 -1.65773566137149
14.0090341213747 -1.89949808435559
15.5938876841942 -2.11178843416667
17.2094102326019 -2.29235928842034
18.8429559937437 -2.40917668452858
20.4780591627055 -2.5268320613002
22.124128436092 -2.57744565609738
23.7774897177226 -2.65382649234217
25.4365672090843 -2.72763769524462
27.1057289198611 -2.65594507538999
};

\addplot [semithick, red]
table {%
-1.04367089271545 -1.96579730510712
-1.04367089271545 -1.96579730510712
-1.04367089271545 -1.96579730510712
-1.04367089271545 -1.96579730510712
-1.04367089271545 -1.96579730510712
0.953486144542694 -1.85942542552948
2.95064330101013 -1.75305342674255
4.94780015945435 -1.64668154716492
6.94495725631714 -1.54030966758728
8.94211483001709 -1.43393778800964
10.9392709732056 -1.32756578922272
12.9364280700684 -1.22119390964508
14.9335851669312 -1.11482203006744
16.9307422637939 -1.00845003128052
18.9279003143311 -0.902078151702881
20.9250564575195 -0.795706212520599
22.9222145080566 -0.689334332942963
24.9193706512451 -0.582962393760681
26.9165287017822 -0.476590484380722
28.9136848449707 -0.370218545198441
30.9108772277832 -0.263936430215836
32.9080467224121 -0.157777830958366
34.9052124023438 -0.0516192391514778
36.9023818969727 0.0545393563807011
38.8995513916016 0.16069795191288
};
\addplot [semithick, red, forget plot]
table {%
-0.830959498882294 -5.9601263999939
-0.830959498882294 -5.9601263999939
-0.830959498882294 -5.9601263999939
-0.830959498882294 -5.9601263999939
-0.830959498882294 -5.9601263999939
1.16619753837585 -5.85375452041626
3.16335463523865 -5.74738264083862
5.16051149368286 -5.64101028442383
7.15766859054565 -5.53463840484619
9.15482616424561 -5.42826652526855
11.1519832611084 -5.32189464569092
13.1491403579712 -5.21552276611328
15.146297454834 -5.10915088653564
17.1434535980225 -5.00277900695801
19.1406116485596 -4.89640712738037
21.137767791748 -4.79003524780273
23.1349258422852 -4.6836633682251
25.1320819854736 -4.57729148864746
27.1292400360107 -4.47091960906982
29.1263961791992 -4.36454725265503
31.123161315918 -4.25828790664673
33.1203308105469 -4.15212965011597
35.1175003051758 -4.04597091674805
37.1146659851074 -3.93981218338013
39.1118354797363 -3.83365368843079
};

\end{axis}

\end{tikzpicture}
		\vspace{-0.4cm}
		\caption{}
		\label{Robutsness1}
		
	\end{subfigure}
	\begin{subfigure}[b]{0.22\textwidth}
		
		\centering
\begin{tikzpicture}

\definecolor{darkgray176}{RGB}{176,176,176}
\definecolor{goldenrod1911910}{RGB}{191,191,0}
\definecolor{green01270}{RGB}{0,127,0}
\definecolor{lightgray204}{RGB}{204,204,204}

\begin{axis}[
	width=\textwidth,
xlabel shift = -3 pt,
ylabel shift = -3 pt,
legend cell align={left},
legend style={
	fill opacity=0.8,
	draw opacity=1,
	text opacity=1,
	at={(0.5,1.335)},
	anchor=north west,
	draw=lightgray204,
	legend columns=-1,
	axis equal
}, 
x grid style={darkgray176},
xlabel={$x \left[ \si{\metre}\right] $ },
xmin=-22, xmax=22,
y grid style={darkgray176},
ylabel={$y \left[ \si{\metre} \right] $ },
ymin=-22, ymax=22,
]
\draw[draw=blue] (axis cs:-1.47050499916077,0) circle (1.4);
\draw[draw=blue] (axis cs:-0.490168333053589,0) circle (1.4);
\draw[draw=blue] (axis cs:0.490168333053589,0) circle (1.4);
\draw[draw=blue] (axis cs:1.47050499916077,0) circle (1.4);

\addplot [semithick, green01270]
table {%
0 0
1.26520859756644 0.0010083939702437
2.52729161134886 -0.000691659727301908
3.78887187141789 0.00598693582300964
5.05055750282332 0.012035882951721
6.31228642634553 0.0174124189750733
7.57320587638792 0.0242137677044401
8.83421906772943 0.0315216922094468
10.0953653499995 0.0387389654504714
11.3568808178765 0.0476494653297275
12.6193556311196 0.0552077751306132
13.882163542211 0.0636687251832862
15.1452962414197 0.0686268947570342
16.4080603200711 0.0707825127231318
17.6698019537302 0.0755249442418569
18.9300279881817 0.0765982419670224
20.1903950790193 0.0822413052573979
21.4505740802151 0.0857351292150911
22.7108765931159 0.0880828236220709
23.9712057851857 0.0856798139486473
};

\addplot [semithick, red]
table {%
-9.27908420562744 1.83256459236145
-7.27914953231812 1.84869682788849
-5.27921438217163 1.86482894420624
-3.27927947044373 1.88096106052399
-1.27934467792511 1.89709317684174
0.720590233802795 1.91322529315948
2.72052526473999 1.92935740947723
4.72045993804932 1.94548952579498
6.7203950881958 1.96162164211273
8.72033023834229 1.97775387763977
10.7202644348145 1.99388599395752
12.7201995849609 2.01001811027527
14.7201347351074 2.02615022659302
16.7200698852539 2.04228234291077
18.7200050354004 2.05841445922852
20.7199401855469 2.07454657554626
22.7198581695557 2.09067869186401
24.7197933197021 2.10681080818176
26.7197284698486 2.12294292449951
28.7196636199951 2.13907504081726
30.7195987701416 2.15520715713501
32.7195320129395 2.17133927345276
34.7194671630859 2.18747138977051
36.7194023132324 2.20360350608826
38.7193374633789 2.21973562240601
};
\addplot [semithick, red, forget plot]
table {%
-9.24681949615479 -2.16730523109436
-7.24688482284546 -2.15117311477661
-5.24695014953613 -2.13504099845886
-3.24701523780823 -2.11890888214111
-1.24708032608032 -2.10277676582336
0.752854585647583 -2.08664464950562
2.75278949737549 -2.07051253318787
4.75272417068481 -2.05438041687012
6.7526593208313 -2.03824830055237
8.75259399414062 -2.02211594581604
10.7525291442871 -2.00598382949829
12.7524642944336 -1.98985183238983
14.7523984909058 -1.97371971607208
16.7523345947266 -1.95758748054504
18.7522678375244 -1.94145536422729
20.7522029876709 -1.92532324790955
22.7521228790283 -1.90919125080109
24.7520580291748 -1.89305913448334
26.7519931793213 -1.87692701816559
28.7519283294678 -1.86079490184784
30.7518634796143 -1.84466278553009
32.7517967224121 -1.82853066921234
34.7517318725586 -1.8123984336853
36.7516670227051 -1.79626631736755
38.7516021728516 -1.7801342010498
};

\end{axis}

\end{tikzpicture}
		\vspace{-0.4cm}
		\caption{}
		\label{Robutsness2}
		
	\end{subfigure}

		\caption{Experiment, in which the SDV is spawned in an infeasible OOD state. (a) Camera image of the scene. (b) Initial infeasible configuration at $t=0\si{\second}$. (c) DCIL successfully leads the vehicle back to the right lane at $t=4\si{\second}$.}
		\label{figg:Robutsness_main}
		\vspace{-0.4cm}
\end{figure}
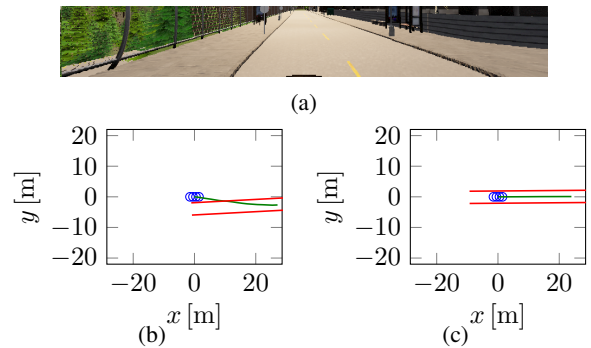

Let us consider question \textit{Q2}. In the CARLA experiments, the bicycle model \cite{Cui2020} is an approximation of the real vehicle dynamics. However, it serves as an \textit{inductive bias}, simplifying the learning process, and enhancing generalization capabilities, as shown by the closed-loop results in Table \ref{tbl:SDE}. To further answer \textit{Q2}, this work conducts another experiment, in which the SDV is spawned onto the wrong lane (Fig. \ref{Robutsness1}). Note that the SDV never encountered such a situation during training, and hence this initial state is entirely outside the manifold of the training data. That situation could occur due to disturbances during driving or because a parked vehicle blocks the lane. Here, some hard constraint methods as \cite{Roesmann2021} provide no solution at all, as the initial state is already infeasible. However, DCIL is robust w.r.t. incorrect specifications, softens the constraints and leads the vehicle back onto the right lane (Fig. \ref{Robutsness2}).

\subsection{Runtime}
The experiments use a AMD Ryzen 9 5900X and a Nvidia RTX 3090.  Our non-optimized python implementation takes on average $20.02 \, \si{\milli\second}$ in the MRE and $119.22\,\si{\milli\second}$ in the SDE. The runtime for the pure IL (SDE) is  $37.20\,\si{\milli\second}$.


\subsection{Discussion, Limitations and Future Work}
\label{Discussion}
This section discusses the limitations of the presented work and gives an outlook on possible future directions. 
First, many active constraints make the loss landscape challenging to optimize and the procedure can be trapped in local minima. We observed that results of SL and DCIL get worse (SDE: both methods' driving score drops by about 15\,\%) using the described squared loss of \cite{Donti}. That can be explained by the fact that the squared loss is more sensitive to outliers and harder to optimize during training, especially in the SDE. Hence we decided to use the non-squared loss of the official implementation \cite{Donti} as mentioned in Section \ref{TrainingInference}.
Second, we marginalize over agents and plan the constrained uni-modal motion of a single vehicle. Future work should extend the approach to constrained joint planning of multi-modal futures with multiple agents, similar to \cite{Diehl2023Icml}, and evaluate the resulting traffic simulation with real-world data \cite{montali2023waymo}.

\section{Conclusion}
This work combined ideas from IL and optimal control for motion planning and control. It accounts for constraints using a differentiable completion and correction procedure. The experiments revealed that our approach outperforms multiple baselines in one mobile robot and one automated driving environment, and can deal with infeasible initial states.


\bibliographystyle{IEEEtran}
\bibliography{IROSW_2023_DCIL_references}

\end{document}